\documentclass[10pt,twocolumn,letterpaper]{article}

\usepackage{cvpr}              

\usepackage{tabularray}
\usepackage{graphicx}
\usepackage{amsmath}
\usepackage{amssymb}
\usepackage{mathrsfs}
\usepackage{array}
\usepackage{pifont}
\usepackage{booktabs}
\usepackage{multirow}
\usepackage{arydshln}
\usepackage{appendix}

\usepackage{algorithm}
\usepackage{algorithmicx}
\usepackage{algpseudocode}

\usepackage{xcolor}
\definecolor{cvprblue}{rgb}{0.21,0.49,0.74}
\usepackage[pagebackref,breaklinks,colorlinks,allcolors=cvprblue]{hyperref}

\usepackage[capitalize]{cleveref}
\crefname{section}{Sec.}{Secs.}
\Crefname{section}{Section}{Sections}
\Crefname{table}{Table}{Tables}
\crefname{table}{Tab.}{Tabs.}

\def\cvprPaperID{0} 
\def\confName{CVPR}
\def\confYear{2025}

\begin{document}

\title{Learning Transformation-Isomorphic Latent Space for \\ Accurate Hand Pose Estimation}  

\author{Kaiwen Ren, Lei Hu, Zhiheng Zhang, Yongjing Ye, Shihong Xia\\
University of Chinese Academic of Science, Institute of Computing Technology, CAS\\
Beijing, China\\
{\tt\small \{renkaiwen23s, hulei19z, zhangzhiheng20g, yeyongjing, xsh\}@ict.ac.cn}
}
\maketitle

\begin{abstract}

Vision-based regression tasks, such as hand pose estimation, have achieved higher accuracy and faster convergence through representation learning. However, existing representation learning methods often encounter the following issues: the high semantic level of features extracted from images is inadequate for regressing low-level information, and the extracted features include task-irrelevant information, reducing their compactness and interfering with regression tasks. To address these challenges, we propose {\rm\textbf{TI-Net}}, a highly versatile visual \textbf{Net}work backbone designed to construct a \textbf{T}ransformation \textbf{I}somorphic latent space. Specifically, we employ linear transformations to model geometric transformations in the latent space and ensure that {\rm TI-Net} aligns them with those in the image space. This ensures that the latent features capture compact, low-level information beneficial for pose estimation tasks. We evaluated {\rm TI-Net} on the hand pose estimation task to demonstrate the network's superiority. On the DexYCB dataset, {\rm TI-Net} achieved a 10\% improvement in the PA-MPJPE metric compared to specialized state-of-the-art (SOTA) hand pose estimation methods. Our code will be released in the future.

\end{abstract}


\begin{figure}[ht]
    \centering
    \includegraphics[width=\linewidth]{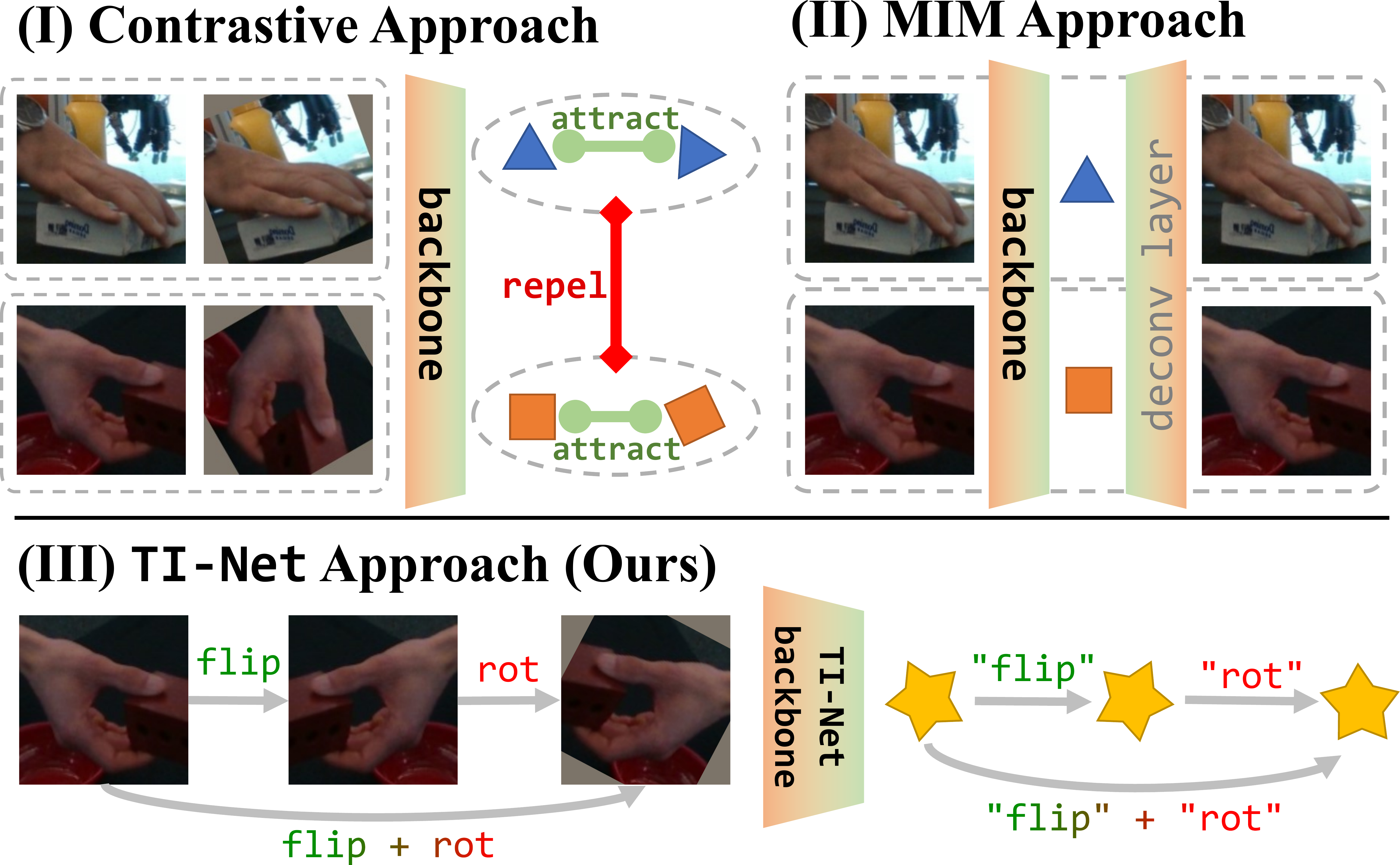}
    \caption{In the pretraining phase, (I) Contrastive learning approaches attract positive pairs and repel negative pairs.\cite{chen_simclr_2020,le-khac_contrastive_2020} (II) Masked image modeling approach reconstructs the image from the embedding of the original one.\cite{he_masked_2021,li_masked_2024} (III) TI-Net ensures that the transformation relationships in the image space also hold in the latent space, as does the combined result of transformations. We refer to this property as ``transformation isomorphism.''}
    \label{img:compare}
\end{figure}

\section{Introduction}
\label{sec:intro}

\begin{figure*}[th]
    \centering
    \includegraphics[width=\linewidth]{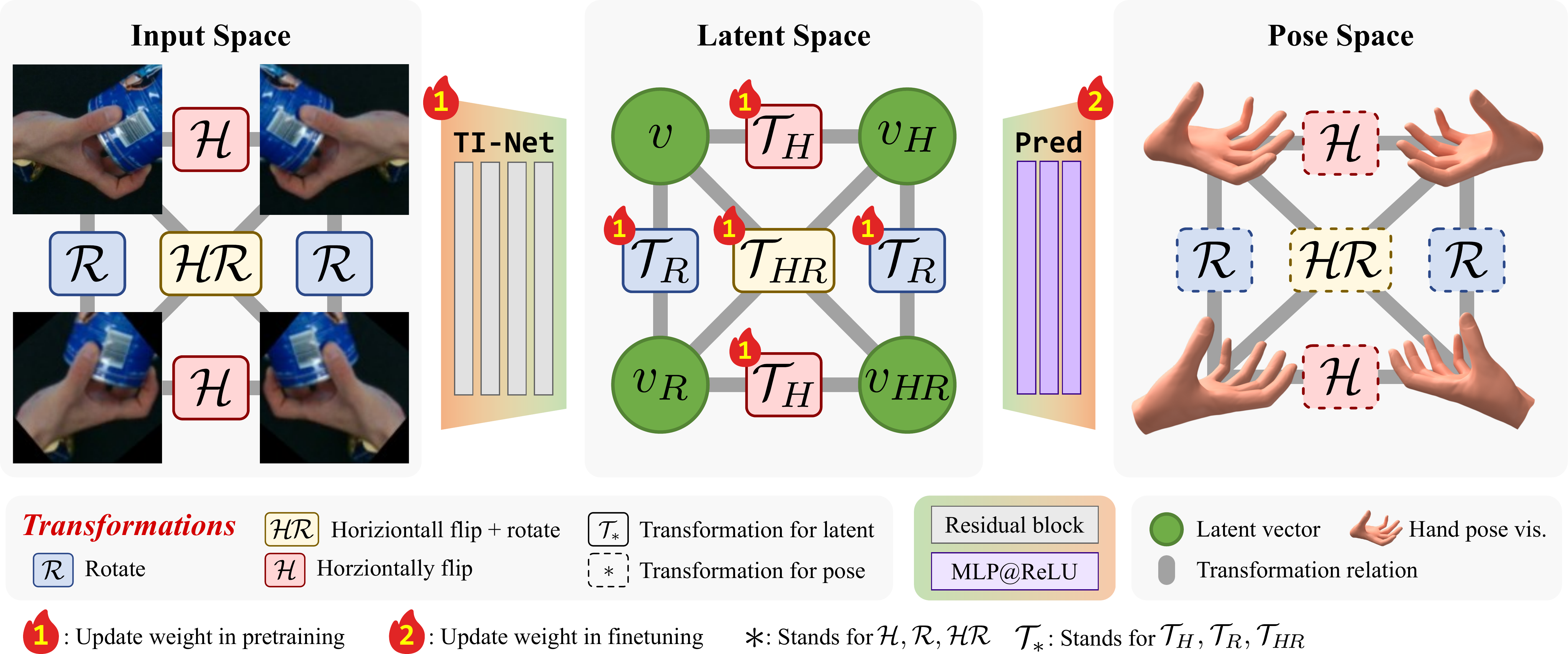}
    \caption{Overview of transformation isomorphism. \textbf{Left}: The relationships among three transformations in the image space: horizontal flip, rotation, and the horizontal flip + rotation. Any two of these transformations can be composed to form another transformation, and the rotation inherently includes the identity transformation. \textbf{Right}: In the pose space, there are three transformations that correspond exactly to the three transformations in the image space, and they satisfy the same combination rules. We refer to this perfect correspondence as transformation isomorphism. \textbf{Center}: TI-Net ensures that there exists transformations in latent space that correspond to the ones in image space. Due to the equivalence property of the isomorphism, the transformations in the latent space also correspond to those in the pose space, and satisfying the same combination rules.}
    \label{img:overview}
\end{figure*}

Hand pose estimation is a crucial component of human-computer interaction systems, serving as a foundational technology that enables natural and intuitive communication between users and machines through the interpretation of dynamic hand gestures. Vision-based hand pose estimation methods have been widely adopted due to their convenience in AR/VR, image and video understanding, and skeleton-based action recognition. 

The primary challenges in estimating hand pose from single-view RGB images include severe self-occlusion and pose diversity. The hand's finer and more complex structure, compared to the human body, increases the likelihood of mutual occlusions among various parts and contributes to a broader range of pose variations. Mainstream hand pose estimation methods can be broadly categorized into two types: task-specific architectures and representation learning approaches. Task-specific architectures involve specially designed networks and branches that directly address the complexities of hand pose estimation, such as creating specialized network structures and utilizing tailored training data to improve accuracy \cite{Yu_2023_CVPR,Ren_2023_ICCV,Li_2022_CVPR,Hampali_2022_CVPR_Kypt_Trans,Zhang_2021_ICCV}. However, these methods typically require a substantial amount of supervised data, which is often costly and labor-intensive to acquire. 

Representation learning approaches \cite{hrnet_sun2019deep,xiao2018simple,xu2022vitpose,xu2022vitpose+,spurr_peclr_2022,lin_pre-training_2024}  first learn features from a large amount of unlabeled data, enabling faster convergence and preventing overfitting to specific datasets. Subsequently, these methods regress the pose from the learned features during the fine-tuning phase. As illustrated in \cref{img:compare}, representation learning methods can be categorized into Masked Image Modeling(MIM)-based and contrastive learning-based approaches \cite{li_masked_2024}. The features extracted by such models may contain rich information about texture, lighting, and color, focusing on image reconstruction rather than pose estimation. Contrastive learning-based methods extract task-relevant features by ensuring that positive sample pairs yield similar feature representations, achieving notable success in high-level tasks such as classification and understanding \cite{HU2024128645,technologies9010002}.  However, since poses are sensitive to image transformations, the extracted features are less suitable for hand pose estimation.

We observe a significant relationship, termed \emph{transformation isomorphism}, between image space and pose space. Transformations applied to the image correspond to those in the pose, as illustrated in \cref{img:overview}. For instance, applying a horizontal flip to the image makes the hand pose in the flipped image equivalent to the flipped version of the original pose. Furthermore, the algebraic structure of transformations in pose space mirrors that in image space, establishing an one-to-one correspondence.

Based on the above observation, we propose \textbf{TI-Net} (``Ti'' stands for \textbf{t}ransformation \textbf{i}somorphism) to embed images into a latent space that maintains a transformation isomorphism relationship with both the image and pose spaces, as depicted in \cref{img:compare}. Because geometric transformations cannot be directly defined in the latent space, it is crucial to organize the latent space to establish an isomorphism with the image space, thereby necessitating the definition of geometric transformations within the latent space. To achieve this, we utilize auxiliary linear transformations to model latent transformations aligned with those in the image space, updating them alongside TI-Net during pre-training. In the fine-tuning stage, these auxiliaries are discarded, enabling TI-Net's seamless integration into existing frameworks.  Therefore, TI-Net not only enhances pose estimation accuracy but also leverages vast unlabeled data to improve generalization capabilities. Our main contributions are as follows:
\begin{itemize}
    \item We present TI-Net to effectively capture the transformation isomorphism relationships among input space, latent space, and target space, which can enhance pose estimation performance. 
    \item TI-Net can be seamlessly integrated into other pose estimation models by replacing the classic vision backbone with ours, requiring minimal modifications. 
    \item We contend that in representation learning for regression tasks, transformation-consistent features are more effective at accurately capturing task-relevant information compared to transformation-invariant features. 
    \item Our proposed method achieves SOTA performance on the pose estimation task. On the DexYCB dataset, it improves PA-MPJPE by approximately $10\%$ compared to the specially designed architecture \cite{Zhou_2024_CVPR}, and achieves an improvement of 1.49mm in MPJPE compared to the SOTA representation learning-based pose estimation method\cite{spurr_peclr_2022}. 
\end{itemize}

\section{Related works}
\label{sec:related}

\subsection{Hand pose estimation}

Mainstream research in hand pose estimation concentrates on estimating hand poses from single-view RGB images or videos \cite{mmpose2020,Lugaresi2019MediaPipeAF,xu2022vitpose,xu2022vitpose+,Hampali_2022_CVPR_Kypt_Trans,10.1109/TPAMI.2023.3247907,10655481,10.1007/978-3-031-20068-7_22,moon_bringing_2023,blur_oh2023recovering3dhandmesh,ego_Prakash2024Hands,lin_pre-training_2024,spurr_peclr_2022}. ViTPose\cite{xu2022vitpose,xu2022vitpose+} successfully integrates Vision Transformer\cite{vit_dosovitskiy2021an} into the pose estimation domain, achieving real-time performance and high precision in 2D pose estimation. Furthermore, the application of advanced architectures such as HRNet\cite{hrnet_sun2019deep,mmpose2020} in 3D hand pose estimation has resulted in significant progress.

A substantial body of work addresses specific challenges in hand pose estimation. For instance, InterWild\cite{moon_bringing_2023} targets the estimation of inter-hand poses in in-the-wild scenarios, \cite{blur_oh2023recovering3dhandmesh} tackles pose estimation in blurred settings, and \cite{ego_Prakash2024Hands} emphasizes first-person hand pose estimation. Concurrently, recent methodologies have emerged that utilize self-supervised learning for hand pose estimation tasks. For example, \cite{lin_pre-training_2024} adopts a 2D pose detector for retrieving similar pose images, allowing the backbone to focus on extracting pose-related features. PeCLR\cite{spurr_peclr_2022} constructs positive samples through geometric transformations, yielding higher accuracy compared to the traditional contrastive learning method for images\cite{chen_simclr_2020}. Other approaches exploit the projection relationship between 3D poses and 2D poses, training networks for pose estimation via self-supervised learning \cite{10.1109/TPAMI.2023.3247907,epipolartransformers,zheng2023hamuco,spurr_peclr_2022}.

Our work, in contrast, adopts a data-driven approach utilizing a purely classical architecture, ResNet\cite{He_2016_CVPR}. By exploiting the isomorphism between input space and latent space, we achieve state-of-the-art (SOTA) performance without employing task-specific pruning designs.

\subsection{Representation learning}

Representation learning seeks to automatically extract valuable latent or representations from raw data. This process allows models to transform complex, high-dimensional data---such as images, text, or audio---into low-dimensional, compact representations that can be effectively utilized for tasks including classification, regression, and clustering. According to \cite{li_masked_2024}, mainstream representation learning methods can be broadly classified into masked image modeling (MIM) methods and contrastive learning methods.

MIM methods enable models to learn the relationships among different regions of an image by reconstructing masked areas based on unmasked regions \cite{he_masked_2021,zhou_ibot_2022,xie2021simmim,Xie_2023_CVPR,Chang_2022_CVPR,Wang_2023_CVPR}. MAE\cite{he_masked_2021} synergizes masked image modeling with ViT \cite{vit_dosovitskiy2021an} to facilitate efficient masked pretraining. Building on the approach established in \cite{he_masked_2021}, numerous representation learning strategies have emerged using masked models. For instance, iBOT \cite{zhou_ibot_2022} incorporates contrastive learning into MIM to capture high-level semantic features, while P-STMO \cite{avidan_p-stmo_2022} extends masked modeling from image space to pose space to learn pose sequence priors.

Contrastive learning methods enable models to learn features beneficial for downstream tasks by bringing closer the representations for positive samples with similar properties \cite{chen_simclr_2020,wang_dense_2021,Mo_2023_WACV,Zheng_2021_ICCV}. \cite{chen_simclr_2020} proposes a foundational framework for contrastive learning within image domain, establishing a baseline for visual contrastive learning. \cite{Zheng_2021_ICCV} treats images belonging to the same class, rather than merely different transformations of the same image, as positive samples for contrastive learning, thus enhancing the model's ability to extract high-level semantic features. Furthermore, \cite{10.1007/978-3-031-19772-7_3} employs 2D poses in contrastive learning, thereby aligning it more effectively with pose estimation tasks. \cite{wang_dense_2021} applies contrastive learning to dense prediction tasks, achieving improvements in object detection, semantic segmentation, and instance segmentation.

\subsection{Learning invariant attributes}

Contrastive learning-based representation learning frameworks generally incorporate a set of transformations applied to the input data, aimed at ensuring that the resulting latent capture information highly relevant to the task. Taking image input as an example, in a contrastive learning framework for high-level tasks (\eg, classification, REID, understanding), this set of transformations often consists of low level transformations like rotations, cropping, scaling, color jittering, noise addition and others. These transformations are crafted to preserve the high-level semantic information contained within the images. Consequently, we expect the backbone to extract \emph{invariant} latent with respect to such transformations under these training frameworks.

Research by \cite{singh_explainable_2024} advocates enabling the network to estimate rotation angles directly from images, while \cite{feng_self-supervised_2019} suggests that the network extract directionally and rotation-invariant features simultaneously. These approaches allow the network to extract high-level semantic information that invariant to rotation, thereby enhancing classification performance. HandCLR \cite{lin_pre-training_2024}, which targeting the pose estimation backbone, introduces the concept of constructing positive samples by retrieving images with analogous poses using a 2D pose detector. This strategy allows the latent extraction network to capture more critical pose information. The aforementioned works concentrating on extracting \textit{transformation invariant} latent, extracting \textit{similar} latent for images subject to different augmentations.

PeCLR\cite{spurr_peclr_2022}, contrarily to \cite{lin_pre-training_2024, singh_explainable_2024, su_ri-mae_2024, feng_self-supervised_2019}, posits that augmentations applied to the image should distinctly affect the corresponding latent representations, resulting in \textit{different} latent for images with varying augmentations. We propose that these different latent should maintain the same structure as images under pre-defined transformations. As shown in (III) of \cref{img:compare}, latent and image compel to same structure with respect to ``flip'', ``rot'' and ``flip+rot''. We refer to this property on latent as \textit{transformation consistency}.

Transformation-consistent latent are capable of better capturing the structure of the hand pose space compared to transformation-invariant features, owing to the transformation-consistency relationship between the image and pose space. By enforcing transformation invariance on latent, the vision backbone inadvertently discard pose-relevant information during the pretraining, leading to inferior estimation accuracy. TI-Net addresses this issue by enforcing transformation isomorphism constraints through the integration of masked image modeling (MIM) and contrastive learning, thereby producing features imbued with transformation consistency and achieving improved accuracy.

\section{Method}
\label{sec:method}

In this section, we introduced the architecture and training framework for TI-Net. Our main contributions focus on pretraining TI-Net to obtain transformation isomorphism. While in finetuning stage we introduce 3D annotated hand pose datasets\cite{Moon_2020_ECCV_InterHand2.6M,chao:cvpr2021_dexycb} to condect supervised learning.

In \cref{sec:method-overview} we detail the idea of transformation isomorphism. Then in \cref{sec:learning-ti-latent} we depict the construction and training framework for TI-Net. Finally in \cref{sec:rotation-embedding} we explain how we encode the rotation parameter into latent transformations. 

\subsection{Modeling}
\label{sec:method-overview}

As shown in \cref{img:overview}, \textit{transformation isomorphism} naturally holds between image space and pose space, with corresponding transformations in image and target space obeying same transformation property with respect to elements in image and target space. It is crucial to notice two important prerequisites of transformation isomorphism:
\begin{enumerate}
    \item Correspondence: there is one-to-one correspondence for transformations in two spaces. The corresponding transformations often share the same semantic.
    \item Consistency: corresponding transformations have follow the same properties, i.e., $f$ and $g$ are corresponding transformation, if $f$ is idempotent then $g$ should also be idempotent.
\end{enumerate}

In our scenario of hand pose estimation, image space and pose space are in transformation isomorphism, with horizontal flipping, rotation, and horizontal flipping+rotation transformation shown in \cref{img:overview} in red, blue and yellow boxes. We denoted transformations in image and pose space as $\mathsf G_{\mathbb I}$ and $\mathsf G_{\mathbb T}$, the isomorphism is denoted as $\mathsf G_{\mathbb I} \cong \mathsf G_{\mathbb T}$.

It is intuitive to hypothesis that will it be possible to organize the latent space $\mathbb L$ and construct transformations for it so that the latent space is transformation isomorphism to image space $\mathsf G_{\mathbb I} \cong \mathsf G_{\mathbb L}$. And the transformation isomorphism among input, latent and target space will follow instantly, termed as $\mathsf G_{\mathbb I} \cong \mathsf G_{\mathbb L} \cong \mathsf G_{\mathbb T}$.

Organizing latent space is the task of TI-Net, the construction of transformations in latent space is a remaining yet important part. In our case, three simple but generic transformations are introduced for image space: $\mathsf G_{\mathbb I}=\{\mathcal H, \mathcal R, \mathcal {HR}\}$, meaning horizontal flipping, rotation, and horizontal flipping+rotation transformation respectively as visualized in \cref{img:overview}. By hypothesis, the corresponding transformations in latent space denoted as $\mathsf G_{\mathbb L}=\{\mathcal T_H, \mathcal T_R, \mathcal T_{HR}\}$ should follow correspondence and consistency prerequisites.

Now, our goal is clear: (1) to find the specific latent transformations $\mathcal T_H, \mathcal T_R, \mathcal T_{HR}$ in latent space $\mathbb L$ that are correspond and consistent to $\mathcal H, \mathcal R, \mathcal {HR}$ in image space $\mathbb I$, (2) to train TI-Net encoding images into latent space where consistency prerequisite of transformations holds.

\begin{table}[!t]
    \centering
    \belowrulesep=0pt\aboverulesep=0pt
    \renewcommand{\arraystretch}{1.2}
    \begin{tabular}{l|lll}
    \toprule
    \multicolumn{1}{c|}{$\circ$} & $\mathcal{H}$   & $\mathcal{R}_{\beta_1}$  & $\mathcal{HR}_{\beta_2}$ \\ \midrule
    $\mathcal{H}$             & $\mathcal{R}_0$ & $\mathcal{HR}_{\beta_1}$ & $\mathcal{R}_{\beta_2}$ \\
    $\mathcal{R}_{\alpha_1}$  & $\mathcal{HR}_{-\alpha_1}$ & $\mathcal{R}_{\alpha_1+\beta_1}$ & $\mathcal{HR}_{-\alpha_1+\beta_2}$ \\
    $\mathcal{HR}_{\alpha_2}$ & $\mathcal{R}_{-\alpha_2}$ & $\mathcal{HR}_{\alpha_2+\beta_1}$ & $\mathcal{R}_{-\alpha_2+\beta_2}$ \\
    \bottomrule
    \end{tabular}
    \caption{Combination results for transformation set incorporating rotations. Note that communicative property is no longer held in this case.}
    \label{tab:complete_transform}
\end{table}
\subsection{Learning transformation isomorphism latent}
\label{sec:learning-ti-latent}


Our objective is to train a vision backbone $\mathcal E_\theta: \mathbb I \to \mathbb L$, such that the transformation set $\mathsf G_{\mathbb I}$ and the corresponding transformation group $\mathsf G_{\mathbb L}$ in the latent space satisfy $\mathsf G_{\mathbb I} \cong_{\mathsf G} \mathsf G_{\mathbb L}$. Such $\mathsf G_{\mathbb I}$ forms a group, the proof can be found in supplement material. To train such vision backbone that extracts transformation isomorphism latent space, we introduce three lightweight latent transformation networks, $\mathcal T_H^\gamma, \mathcal T_R^\eta, \mathcal T_{HR}^\lambda$ to model $\mathcal T_H, \mathcal T_R, \mathcal T_{HR}$ and guide the isomorphism relation between spaces ($\gamma,\eta,\lambda$ are learnable parameters).

\begin{figure}[!t]
    \centering
    \includegraphics[width=\linewidth]{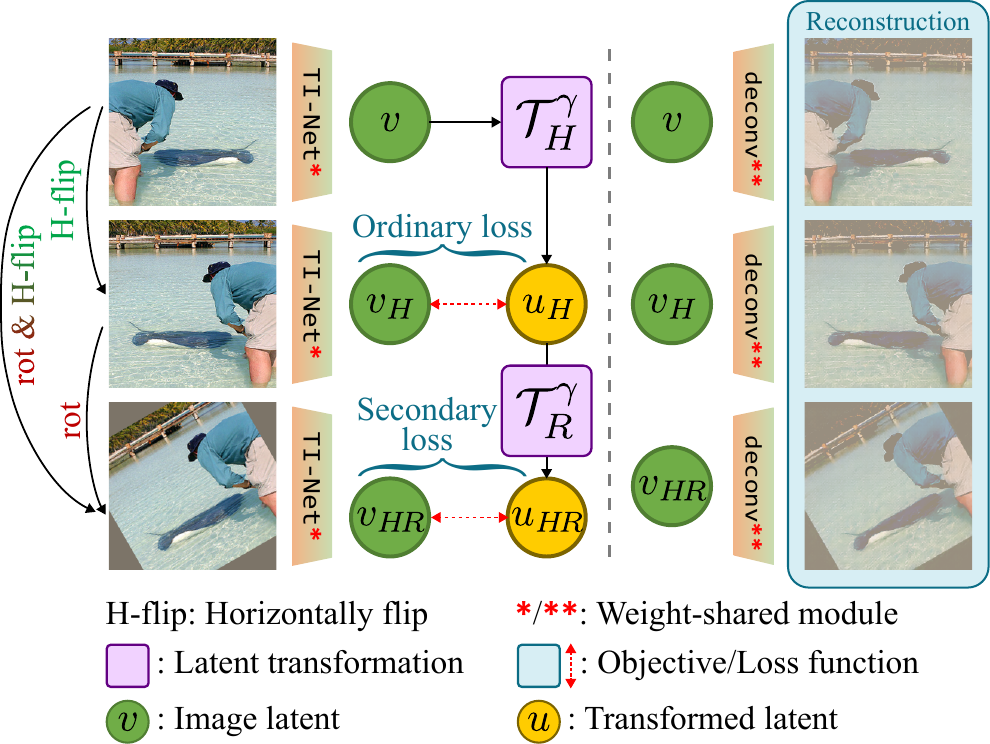}
    \caption{Simplified overview of pretraining phase. Weights of latent transformation are updated jointly with TI-Net. We depict only one ordinary and one secondary constraint here for simplicity.}
    \label{fig:train}
\end{figure}

To mitigate overfitting the latent transformations within the latent space, we use low-degree-of-freedom linear transformations $H, R, B \in \mathbb{R}^{d \times d}$ to construct the latent transformation $\mathcal T_H^\gamma, \mathcal T_R^\eta, \mathcal T_{HR}^\lambda$. Taking the $\mathcal T_H^\gamma$ as an example, it is formulated as $H=E + M_HN_H$, where $\gamma = \{M_H, N_H\}$, $M_H, N_H^\top \in \mathbb{R}^{d \times r}$, and $r \ll d$, $E$ being the identity matrix. This construction not only reduces the risk of overfitting the latent space transformations but also ensures that the transformation is likely to be full-rank, aligning with the property that the flip transformation $\mathcal H$ is a full-rank linear transformation.

We extend the notation of $\mathcal R$ to $\mathcal R_\alpha$, which means rotating the image around its center for $\alpha$ degree counterclockwise and $\mathcal {HR}_\alpha$ as horizontally flipping then rotating the image, so does $\mathcal T_R^\eta(\cdot;\alpha)$ and $\mathcal T_{HR}^\lambda(\cdot;\alpha)$. The combination relations between every pair of elements in $\mathsf G_{\mathbb I}$ are shown in \cref{tab:complete_transform} (proof in supplement material). By transformation isomorphism $\mathsf G_{\mathbb I}\cong_{\mathsf G}\mathsf G_{\mathbb L}$, we assert members of all parameterized latent transformations $\mathsf G_{\mathbb L}=\{\mathcal T_H^\gamma, \mathcal T_R^\eta, \mathcal T_{HR}^\lambda\}$ should also obey the same combination relations. We utilized such obedience to construct our pretraining framework.

The key to learning a transformation isomorphism latent space is the obedience to relations in \cref{tab:complete_transform}. Besides updating vision backbone TI-Net, $\mathcal T_H^\gamma, \mathcal T_R^\eta$ and $\mathcal T_{HR}^\lambda$ should also be updated jointly. We define our loss function as follows:
\begin{equation}
    \mathcal L_{\rm TI} :=\mathcal L_{\rm classic} + w(\mathcal L_{\rm ord} + \mathcal L_{\rm sec}),
\end{equation}
here, $\mathcal L_{\rm classic}$ represents the traditional reconstruction loss, enforcing image reconstruction from the latent of the original and transformed images using auxiliary network $\mathcal D_\Theta^{\rm aux}$ with learnable parameter $\Theta$. $w$ is a weight coefficient, setting to $w = 0.001$. $\mathcal L_{\rm ord}$ and $\mathcal L_{\rm sec}$ corresponds to the loss for the ordinary and secondary constraints:
\begin{equation}
    \left\{
    \begin{aligned}
        &\begin{aligned}
        \mathcal L_{\rm ord} :=\,
            & \Vert \mathcal E_\theta(\mathcal H(I))-\mathcal T_H^\gamma(\mathcal E_\theta(I)) \Vert\, + \\
            & \Vert \mathcal E_\theta(\mathcal R_\alpha(I))-\mathcal T_R^\eta(\mathcal E_\theta(I);\alpha) \Vert\, + \\
            & \Vert \mathcal E_\theta(\mathcal {HR}_\beta(I))-\mathcal T_{HR}^\lambda(\mathcal E_\theta(I);\beta) \Vert ,
        \end{aligned} \\
        &\begin{aligned}
        \mathcal L_{\rm sec} :=\,
            & \Vert \mathcal E_\theta(I)-\mathcal T_H^\gamma (\mathcal T_H^\gamma(\mathcal E_\theta(I))) \Vert\, + \\
            & \Vert \mathcal E_\theta(\mathcal {HR}_\omega(I))-\mathcal T_R^\eta(\mathcal T_H^\gamma(\mathcal E_\theta(I));\omega) \Vert + \cdots,
        \end{aligned}
    \end{aligned}
    \right.
\end{equation}
$\alpha,\beta,\omega$ are randomly sampled from $[0,2\pi]$. The term ``ordinary'' means only 1 latent transformation is applied to the latent, and the term ``secondary'' means 2 latent transformations are applied to the latent one after the other. Theoretically, we can combine arbitrary $N=2,3,\cdots$ latent transformations to get $N$-\textit{th} loss, but it is practically infeasible.

\cref{fig:train} illustrates the training framework with one ordinary constraints $\Vert \mathcal E_\theta(\mathcal H(I))-\mathcal T_H^\gamma(\mathcal E_\theta(I)) \Vert=\mathbf 0$ and one secondary constraint $\Vert \mathcal E_\theta(\mathcal {HR}_\omega(I))-\mathcal T_R^\eta(\mathcal T_H^\gamma(\mathcal E_\theta(I));\omega) \Vert=\mathbf 0$. Readers can refer to \cref{code:pretraining} for complete pretraining procedure.

\begin{algorithm}[!t]
    \small
    \caption{Transformation isomorphism learning}
    \label{code:pretraining}
    \begin{algorithmic}[1]
        \State \textbf{Input:} input image $I$.
        \State \textbf{Output:} $\mathcal L_{\rm TI}$.
        \State $\mathcal L_{\rm ord}=0$ \Comment{Ordinary loss}
        \State Randomly sample $\alpha,\beta\sim\mathcal U(0,2\pi)$
        \State $\mathcal T_R^\eta=\mathcal T_R^\eta(\cdot;\alpha), \mathcal T_{HR}^\lambda=\mathcal T_{HR}^\lambda(\cdot;\beta)$
        \For {$(\mathcal F,\mathcal T^\phi)$ in $\{(\mathcal H,\mathcal T_H^\gamma),(\mathcal R,\mathcal T_R^\eta),(\mathcal {HR},\mathcal T_{HR}^\lambda)\}$}
            \State $\mathcal L_{\rm ord}=\mathcal L_{\rm ord}+
                \Vert
                    \mathcal E_\theta(\mathcal F(I)) - \mathcal T^\phi(\mathcal E_\theta(I))
                \Vert
            $
        \EndFor
        \State $\mathcal L_{\rm sec}=0$ \Comment{Secondary loss}
        \State Randomly sample $\alpha,\beta\sim\mathcal U(0,2\pi)$
        \State $\mathcal T_R^\eta=\mathcal T_R^\eta(\cdot;\alpha), \mathcal T_{HR}^\lambda=\mathcal T_{HR}^\lambda(\cdot;\beta)$
        \For {$(\mathcal F_1,\mathcal T^\phi_1,\mathcal F_2,\mathcal T^\psi_2)$ in $\{(\mathcal H,\mathcal T_H^\gamma),(\mathcal R,\mathcal T_R^\eta),(\mathcal {HR},\mathcal T_{HR}^\lambda)\}^2$}
            \State $\mathcal L_{\rm sec}=\mathcal L_{\rm sec}+
                \Vert
                    \mathcal E_\theta(\mathcal F_1 \circ \mathcal F_2 (I)) -
                    \mathcal T^\phi_2(\mathcal T^\psi_1(\mathcal E_\theta(I)))
                \Vert
            $
        \EndFor
        \State $\mathcal L_{\rm classic}=0$ \Comment{Reconstruction loss}
        \For {$\mathcal F$ in $\{\mathcal H,\mathcal R,\mathcal {HR}\}$}
            \State $\mathcal L_{\rm classic}=\mathcal L_{\rm classic}+\Vert \mathcal D^{\rm aux}_{\Theta}(\mathcal E_\theta(\mathcal F(I))) - \mathcal F(I) \Vert$
        \EndFor
        \State $\mathcal L_{\rm TI}=\mathcal L_{\rm classic}+w(\mathcal L_{\rm ord}+\mathcal L_{\rm sec})$ \Comment{TI loss}
    \end{algorithmic}
\end{algorithm}


\subsection{Rotation embedding}
\label{sec:rotation-embedding}

Different from $\mathcal T_H$, $\mathcal T_{H}(\cdot;\alpha)$ and $\mathcal T_{HR}(\cdot;\alpha)$ are parameterized with continuous number $\alpha$. Therefore, when constructing the corresponding transformations in the latent space, \eg $\mathcal T_R^\eta(\cdot;\alpha)$ and $\mathcal T_{HR}^\lambda(\cdot;\alpha)$, the rotation parameter $\alpha$ needs to be treated as input into the transformation. The specific form of the transformation remains a linear one, but the rotation parameter is embedded and concatenated with the latent features as preprocessing. Taking the rotation transformation $\mathcal T_R^\eta(\cdot;\alpha)$ as an example:
\begin{equation}
    \begin{aligned}
        \mathcal T_R^\eta(v;\alpha)&=v+M_RN_R\cdot \verb|cat|(v,e_r) \\
        e_r &= \mathcal K_\xi (r) \\
        r &= [\cos\alpha,\sin\alpha]^\top
    \end{aligned}
\end{equation}
Here, $\mathcal K_\xi$ represents a rotation vector embedding network shared by $\mathcal T^\eta_R(\cdot;\alpha)$ and $\mathcal T^\lambda_{HR}(\cdot;\alpha)$ with learnable parameters $\xi=\eta\cap\lambda$. It is a two-layer MLP for embedding the two-dimensional rotation direction vector into an $n$-dimensional space. $v$ denotes the latent features of the original image.



\begin{figure*}[!t]
    \centering
    \includegraphics[width=0.9\linewidth]{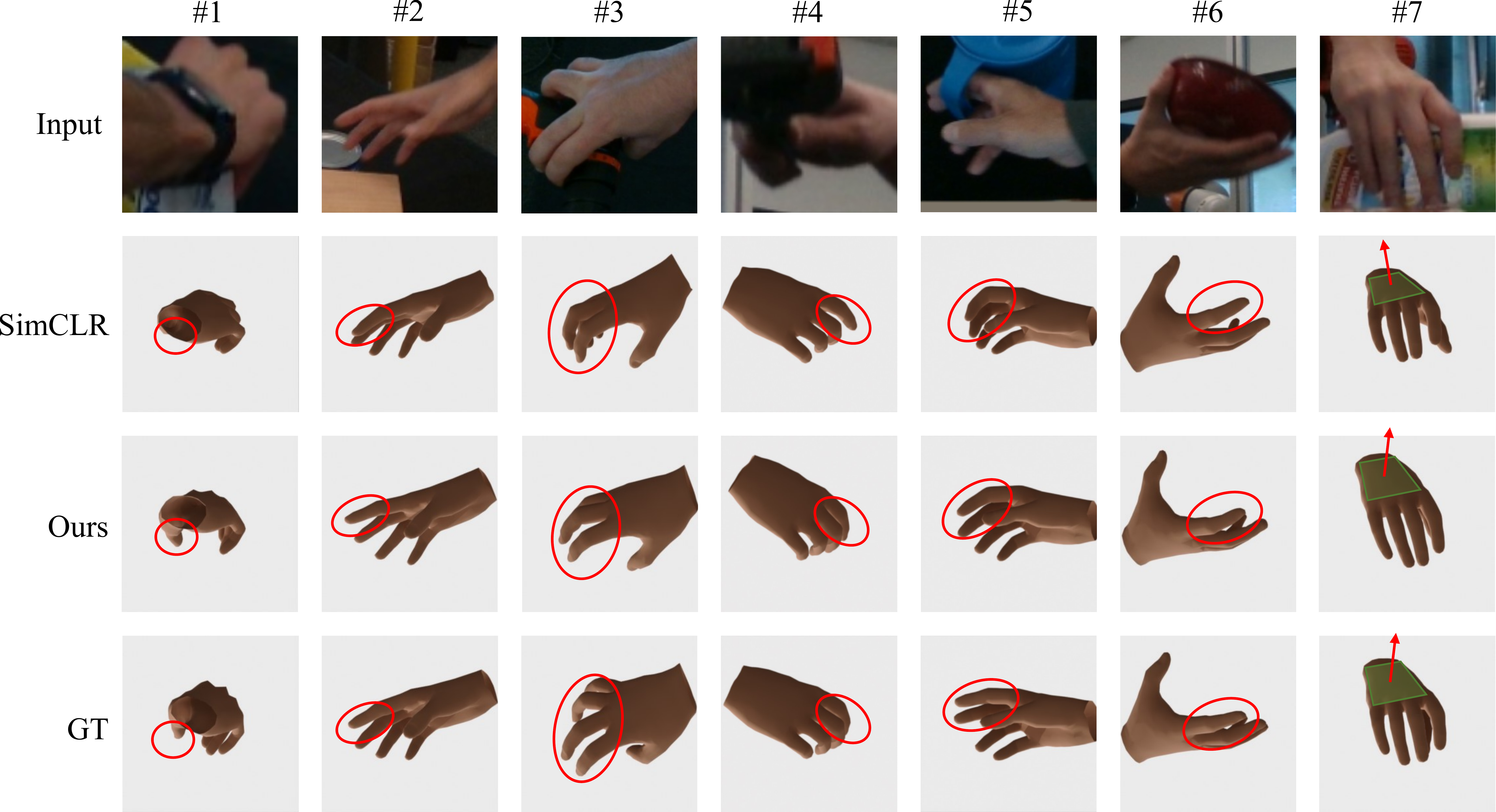}
    \caption{Visualization comparison between our approach and SimCLR\cite{chen_simclr_2020} on DexYCB\cite{chao:cvpr2021_dexycb} Our method exhibits better accuracy under occlusion scenes. TI-Net and SimCLR\cite{chen_simclr_2020} are finetuned on DexYCB\cite{chao:cvpr2021_dexycb} under the same procedure and meta-parameters. GT standards for ground truth annotations. We adjusted the viewing direction for the best comparison.}
    \label{fig:vis-compare}
\end{figure*}

\section{Experiments}

\subsection{Implementation}

We use ResNet \cite{He_2016_CVPR} as the backbone for pretraining. The network takes $224\times 224$ RGB images as input and outputs an image feature map with the size of $7\times 7\times 2048$. Based on our proposed training framework, we train the model for 50 epochs on the ImageNet-1K dataset using the AdamW optimizer. The batch size is set to 68, and the base learning rate is $1.5\times10^{-3}$. For the finetuning stage, we crop the input to $224\times 224$ as the input for TI-Net and add a 3-layer MLP to estimate the MANO pose parameters $\theta \in \mathbb{R}^{16\times 3}$ from the features.
\subsection{Datasets}

\paragraph{DexYCB}
The DexYCB\cite{chao:cvpr2021_dexycb} dataset is a large-scale benchmark designed for 3D hand pose estimation and hand-object interaction tasks. It contains synchronized RGB-D videos of human hands interacting with 20 YCB objects, providing accurate 3D annotations for both hand joints and object poses. The dataset offers diverse scenarios, including various grasp types and hand-object occlusions. It is suitable for training and evaluating models to understand complex hand movements and interactions in real-world settings.

\paragraph{InterHand2.6M}
The InterHand2.6M\cite{Moon_2020_ECCV_InterHand2.6M} dataset is a large-scale dataset specifically created for 3D hand pose estimation, featuring over 2.6 million annotated hand images. It contains both single-hand and interacting-hand scenarios, captured from multiple camera angles with diverse poses. The dataset provides high-quality 3D annotations of hand joints, making it a valuable resource for training and evaluating models aimed at accurately estimating complex hand poses, including hand-to-hand interactions.

\subsection{Evaluation metrics}

\paragraph{MPJPE}
Mean Per Joint Position Error (MPJPE) measures the average Euclidean distance between the predicted and ground truth 3D joint positions after root alignment. It is defined as:
\begin{equation}
    \text{MPJPE} = \frac{1}{N} \sum_{i=1}^{N} \| \mathbf{P}_i - \mathbf{G}_i \|_2,
\end{equation}
where \( N \) denotes the number of joints, \( \mathbf{P}_i \) represents the predicted position of the \( i \)-th joint, and \( \mathbf{G}_i \) is the corresponding ground truth position. MPJPE captures the overall accuracy of the estimated hand pose by providing a direct measure of the discrepancy between predicted and actual joint locations, with lower MPJPE values indicating better performance. It is particularly effective for assessing the precision of 3D hand pose estimation models.

\paragraph{PA-MPJPE}
Procrustes-Aligned Mean Per Joint Position Error (PA-MPJPE)\cite{Ionescu2014} is an evaluation metric for hand pose estimation that measures the average Euclidean distance between the predicted and ground truth 3D joint positions after Procrustes alignment\cite{Gower1975}. This alignment step removes variations in translation, rotation, and scale, focusing solely on the pose similarity. PA-MPJPE is defined as:
\begin{equation}
    \text{PA-MPJPE} = \frac{1}{N} \sum_{i=1}^{N} \| \hat{\mathbf{P}}_i - \mathbf{G}_i \|_2,
\end{equation}
where \( \hat{\mathbf{P}}_i \) denotes the Procrustes-aligned prediction of the \( i \)-th joint. Lower PA-MPJPE values indicate better alignment with the ground truth, making it a robust metric for assessing pose accuracy.

\subsection{Comparison with representation learning methods}

\paragraph{Quantitative comparison}
We compare the performance of our approach with current mainstream image feature networks that use representation learning for pretraining in the hand pose estimation task, as shown in \cref{tab:representation}.

\begin{table}[!t]
    \centering
    \small
    \begin{tabular}{lllc}
        \toprule
        Method & Pretraining & Backbone & MPJPE \\ \midrule
        \multirow{3}{*}{SimCLR\cite{chen_simclr_2020}} & 100DOH-1M\cite{Shan20} & ResNet-50 & 20.13 \\
         & Ego4D-1M\cite{Grauman_2022_CVPR} & ResNet-50 & 20.22 \\
         & ImageNet-1K\cite{Deng2009} & ResNet-50 & 20.32 \\
        \hline
        \multirow{2}{*}{PeCLR\cite{spurr_peclr_2022}} & 100DOH-1M\cite{Shan20} & ResNet-50 & 18.39 \\
         & Ego4D-1M\cite{Grauman_2022_CVPR} & ResNet-50 & 18.99 \\
        \hline
        \textbf{Ours} & ImageNet-1K\cite{Deng2009} & ResNet-50 & \textbf{16.79} \\
        \bottomrule
    \end{tabular}
    \caption{Comparison of MPJPE results with SOTA representation learning method. The evaluation metrics in the table are referenced from \cite{lin_pre-training_2024}.}
    \label{tab:representation}
\end{table}

Different pretraining tasks and datasets can have varying impacts on pose estimation results. Choi et al. \cite{choi2023rethinking} demonstrated that for pose estimation tasks, using pose images (\eg, SURREAL\cite{Varol_2017_CVPR}) for pretraining yields a backbone that achieves better accuracy compared to ones pretrained on ImageNet-1K\cite{Deng2009}. In \cref{tab:representation}, TI-Net outperforms current SOTA representation methods, even with those pretrained on hand relevant datasets.

\paragraph{Qualitative comparison}
Visualization results are shown in \cref{fig:vis-compare}. Due to transformation isomorphism between image, latent, and pose space, TI-Net achieves a more accurate hand pose as shown in cases \#2, \#6, and \#7, where major parts of the hand are visible in the image. while in occluded cases like \#1, \#3, \#4, and \#5, our approach produces more reasonable poses because the latent TI-Net extracted is more compact and contains more information about the pose, reducing the interference of irrelevant information and alleviating the lack of information in occluded regions.

\begin{figure}
    \centering
    \includegraphics[width=0.8\linewidth]{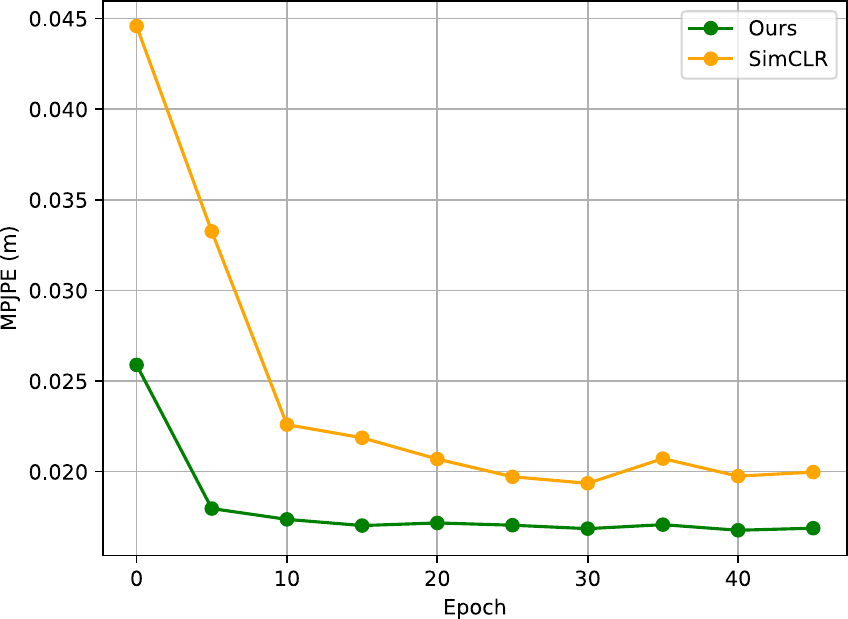}
    \caption{Comparison of MPJPE between our approach and SimCLR\cite{chen_simclr_2020} approach on every epoch on DexYCB\cite{chao:cvpr2021_dexycb}, aligning all training setup the same. Our approach shows faster convergence and more stable training.}
    \label{fig:epoch-mpjpe}
\end{figure}

\paragraph{Training efficiency}
We plot the MPJPE of every five epochs for SimCLR\cite{chen_simclr_2020} and our approach in \cref{fig:epoch-mpjpe}. As shown in the figure, TI-Net presents better training efficiency with faster convergence and more stable decrement. This indicates that SimCLR\cite{chen_simclr_2020} requires more substantial parameter adjustments during finetuning to fit regression tasks effectively. In contrast, TI-Net, by leveraging transformation isomorphism to construct a structure similar to the target space, gains faster training speed.

\subsection{Comparison with other methods}

\paragraph{Evaluation on DexYCB}
We evaluate our model on the large-scale hand-object interaction dataset, DexYCB \cite{chao:cvpr2021_dexycb}. DexYCB contains multi-view hand grasping data, making hand pose estimation more challenging due to the presence of occlusions. We assess both PA-MPJPE and MPJPE metrics, with the results shown in \cref{tab:dexycb}. It is observed that our model achieves the best PA-MPJPE results, even compared to the methods with specifically designed architecture or the ones incorporating temporal information, reaching state-of-the-art performance. This indicates that the transformation isomorphic latent space constructed by our approach has a positive impact, effectively improving the accuracy of hand pose estimation.

\begin{table}[!h]
    \centering
    \small
    \begin{tabular}{lccc}
        \toprule
        Method & S & PA-MPJPE & MPJPE \\ \midrule
        METRO\cite{Zhang_2019_ICCV} & \checkmark & 7.0 & 15.2   \\
        Spurr et al.\cite{Spurr2020} & \checkmark & 6.8 & 17.3  \\
        Liu et al.\cite{Liu_2021_CVPR} & \checkmark & 6.6 & 15.3  \\
        MobRecon\cite{Chen_2022_CVPR} & \checkmark & 6.4 & 14.2  \\
        HandOccNet\cite{Park_2022_CVPR} & \checkmark & 5.8 & 14.0  \\
        H2ONet\cite{Xu_2023_CVPR}${}^\star$ & \checkmark & 5.7 & 14.0   \\
        Zhou et al.\cite{Zhou_2024_CVPR} & \checkmark & 5.5 & \textbf{12.4}  \\
        Deformer\cite{Fu_2023_ICCV}${}^\star$ & \checkmark & \underline{5.2} & \underline{13.6}  \\
        \hline
        \textbf{Ours (TI-Net + 3$\times$MLP)} & \ding{55} & \textbf{4.91} & 16.8  \\
        \bottomrule
    \end{tabular}
    \caption{Comparison results on the DexYCB\cite{chao:cvpr2021_dexycb} dataset. Our method achieves the best PA-MPJPE results, indicating a more precise estimation of local hand poses. ``S'' indicates whether the work utilized a task-specifically designed network architecture for hand pose estimation. Methods remarked with ``$\star$'' incorporates temporal information.}
    \label{tab:dexycb}
\end{table}

\paragraph{Evaluation on InterHand2.6M}
InterHand2.6M is a large-scale dataset collected in the lab setting, with challenging cases of interacting hands. We evaluate TI-Net on InterHand2.6M using PA-MPJPE and MPJPE metrics and compare it with related works, as shown in \cref{tab:ih26m}. TI-Net achieves better results than Keypoint Transformer\cite{Hampali_2022_CVPR_Kypt_Trans} and InterWild\cite{moon_bringing_2023}, demonstrating its capability for accurate pose estimation. However, there is a slight inferiority between TI-Net and IntagHand\cite{Li_2022_CVPR}. We believe there are two main reasons for this phenomenon: (1) IntagHand employs a highly specialized model structure, incorporating modules like the Interacting Attention Graph, in addition to the shared ResNet50 backbone, specifically designed for the dual-hand pose estimation scenario; (2) the more severe hand occlusions and greater similarity between left and right hands in InterHand2.6M, which significantly impact the model and cannot be mitigated by the transformation isomorphic relationship.  

\begin{table}[!t]
    \centering
    \small
    \begin{tabular}{lccc}
        \toprule
        Method & S & PA-MPJPE & MPJPE \\ \midrule
        InterShape\cite{Zhang_2021_ICCV} & \checkmark & - & 13.07 \\
        Keypoint Transformer\cite{Hampali_2022_CVPR_Kypt_Trans} & \checkmark & - & 12.78 \\
        InterWild\cite{moon_bringing_2023} & \checkmark & - & 11.67 \\ 
        IntagHand\cite{Li_2022_CVPR} & \checkmark & - & \textbf{8.79} \\
        \hline
        \textbf{Ours (TI-Net + 3$\times$MLP)} & \ding{55} & 4.47 & \underline{10.34} \\
        \bottomrule
    \end{tabular}
    \caption{Comparison results on the InterHand2.6M \cite{Moon_2020_ECCV_InterHand2.6M} dataset. TI-Net exhibits relatively inferior results compared to SOTA methods. However, it still achieves accurate pose estimation and demonstrates better precision compared to \cite{Zhang_2021_ICCV,Hampali_2022_CVPR_Kypt_Trans,moon_bringing_2023}.}
    \label{tab:ih26m}
\end{table}

\paragraph{Analysis}
We observe that TI-Net achieves excellent PA-MPJPE results, performing better with both the task-specialized methods and the temporal information-enhanced methods on the DexYCB\cite{chao:cvpr2021_dexycb} dataset. We attribute this to the transformation isomorphism relationship, which more accurately describes how joint rotations change in response to geometric transformations of the image. For example, a horizontal flip of the image results in a left-right reversal of the orientation of each hand joint. This implies that the features extracted by TI-Net are more sensitive to local hand features, leading to a more accurate estimation of relative hand poses. Since PA-MPJPE measures the similarity between poses solely, TI-Net achieves better PA-MPJPE scores.


\section{Conclusion}

We propose TI-Net, a network that enables accurate hand pose estimation by constructing transformation isomorphism relationships among the input space, latent space, and pose space. This method is based on existing network architectures and enhances task-relevant vision backbone through the introduction of a novel transformation isomorphism, allowing seamless integration into other pose estimation frameworks without significant modifications to the network structure or training pipeline.

In future work, we plan to extend TI-Net from pose estimation tasks to vision regression tasks and enable more backbone architectures to extract transformation isomorphism relationships, improving the performance and effectiveness of regression tasks in computer vision.


{\small
\bibliographystyle{ieee_fullname}
\bibliography{egbib.bib}
}

\end{document}


\title{Supplementary Material for \\ ``Learning Transformation-Isomorphic Latent Space for \\ Accurate Hand Pose Estimation''}

\author{Kaiwen Ren, Lei Hu, Zhiheng Zhang, Yongjing Ye, Shihong Xia\\
Institute of Computing Technology, CAS\\
Beijing, China\\
{\tt\small \{renkaiwen23s, hulei19z, zhangzhiheng20g, yeyongjing, xsh\}@ict.ac.cn}
}
\maketitle

\section{Prove the property of transformation}

We need to prove that the transformation set \( \mathsf G_{\mathbb I}=\{\mathcal H, \mathcal R_\theta, \mathcal{HR}_\theta \} \) satisfies the definition of group: closure for group combination operation, existence of identity element and existence of inverse for every element. The proof of the combination rule for \( \mathsf{G}_{\mathbb{I}} \) is simultaneously established through the proof of the closure property.

\paragraph{Proof of closure}

Before the formal proof, we first describe the properties of the transformations in \( \mathsf{G}_{\mathbb{I}} \).
\begin{align}
    \mathcal {HR}_\theta &= \mathcal H \circ \mathcal R_\theta, \label{eq1} \\
    \mathcal H \circ \mathcal R_\theta &= \mathcal R_{-\theta} \circ \mathcal H,
\end{align}
for any 2D coordinate $(x,y)$, by transformations $\mathcal H\circ\mathcal R_\theta$ we mean first flipping $(x,y)$ to $(-x,y)$ then rotate for $\theta$ radius to get $(-x\cos\theta-y\sin\theta,-x\sin\theta+y\cos\theta)$. Whereas if we first rotate for $-\theta$ radius we get $(x\cos\theta+y\sin\theta,-x\sin\theta+y\cos\theta)$, then horizontal flip produces $(-x\cos\theta-y\sin\theta,-x\sin\theta+y\cos\theta)$, which is the same as previous result. The proof for Eq.1 is similar.

According to the associative property of matrix multiplication, the same property applies to $\mathsf G_{\mathbb I}$ instantly.

Enumerate all possible pairs of elements in \( \mathsf{G}_{\mathbb{I}} \), and prove that their composition remains within \( \mathsf{G}_{\mathbb{I}} \):
\begin{itemize}
    \item Case 1: \( \mathcal H \circ \mathcal H = \mathcal I = \mathcal R_0 \),
    \item Case 2: \( \mathcal R_\alpha \circ \mathcal R_\beta = \mathcal R_{\alpha + \beta} \),
    \item Case 3: \( \mathcal {HR}_\alpha \circ \mathcal {HR}_\beta = \mathcal H \circ \mathcal R_\alpha \circ \mathcal H \circ \mathcal R_\beta = \mathcal R_{-\alpha}  \circ \mathcal R_\beta = \mathcal R_{-\alpha + \beta} \),
    \item Case 4: \( \mathcal H\circ\mathcal R_\theta=\mathcal{HR}_\theta, \mathcal R_{\theta}\circ\mathcal H=\mathcal H\circ\mathcal R_{-\theta}=\mathcal{HR}_{\theta} \),
    \item Case 5: \( \mathcal H\circ\mathcal {HR}_{\theta}=\mathcal R_{\theta}, \mathcal {HR}_{\theta}\circ\mathcal H=\mathcal R_{-\theta} \),
    \item Case 6: \( \mathcal R_{\alpha}\circ\mathcal {HR}_{\beta} = \mathcal {HR}_{-\alpha+\beta}, \mathcal {HR}_{\beta}\circ\mathcal R_{\alpha}=\mathcal {HR}_{\alpha+\beta} \),
\end{itemize}
in summary, we have completed the proof of closure.

\paragraph{Existence of identity}
\( \mathcal{R}_0 \) is the clearly identity element.

\paragraph{Existence of inverse}
The inverse elements of \( \mathcal{H} \), \( \mathcal{R}_{\theta} \), and \( \mathcal{HR}_{\theta} \) are \( \mathcal{H} \), \( \mathcal{R}_{-\theta} \), and \( \mathcal{HR}_{-\theta} \), respectively.

In conclusion, we have proven that \( \mathsf{G}_{\mathbb{I}} \) is a group.   \(\blacksquare\)
